\documentclass[sigconf]{acmart}

\AtBeginDocument{%
  \providecommand\BibTeX{{%
    \normalfont B\kern-0.5em{\scshape i\kern-0.25em b}\kern-0.8em\TeX}}}

\copyrightyear{2021}
\acmYear{2021}
\setcopyright{acmcopyright}\acmConference[SIGSPATIAL '21]{29th International Conference on Advances in Geographic Information Systems}{November 2--5, 2021}{Beijing, China}
\acmBooktitle{29th International Conference on Advances in Geographic Information Systems (SIGSPATIAL '21), November 2--5, 2021, Beijing, China}
\acmPrice{15.00}
\acmDOI{10.1145/3474717.3483920}
\acmISBN{978-1-4503-8664-7/21/11}

\usepackage{multirow}
\usepackage{booktabs}
\usepackage{graphicx}
\usepackage{color}

\begin{document}

\title{DetectorNet: Transformer-enhanced Spatial Temporal Graph Neural Network for Traffic Prediction}




\author{He Li}
\authornote{Both authors contributed equally to this research.}
\affiliation{%
	\institution{Xidian University}
	\city{Xi'an}
	\country{China}}
\email{heli@xidian.edu.cn}

\author{Shiyu Zhang}
\authornotemark[1]
\affiliation{%
\institution{Xidian University}
\city{Xi'an}
	\country{China}}
\email{sy_zhang@stu.xidian.edu.cn}

\author{Xuejiao Li}
\affiliation{%
	\institution{Xidian University}
	\city{Xi'an}
	\country{China}}
\email{xjli_521@stu.xidian.edu.cn}

\author{Liangcai Su}
\affiliation{%
	\institution{Xidian University}
	\city{Xi'an}
	\country{China}}
\email{suliangcai@stu.xidian.edu.cn}

\author{Hongjie Huang}
\affiliation{%
	\institution{Xidian University}
	\city{Xi'an}
	\country{China}}
\email{huanghongjie@stu.xidian.edu.cn}

\author{Duo Jin}
\affiliation{%
	\institution{Xidian University}
	\city{Xi'an}
	\country{China}}
\email{djin@stu.xidian.edu.cn}
	
\author{Linghao Chen}
\affiliation{%
	\institution{Xidian University}
	\city{Xi'an}
	\country{China}}
\email{lhchen@stu.xidian.edu.cn}

\author{Jianbin Huang}
\authornote{Corresponding author.}
\affiliation{%
    \institution{Xidian University}
	\city{Xi'an}
	\country{China}}
\email{jbhuang@xidian.edu.cn}
    
\author{Jaesoo Yoo}
\affiliation{%
	\institution{Chungbuk National University}
	\city{}
	\country{South Korea}}
\email{yjs@cbnu.ac.kr}

\renewcommand{\shortauthors}{He Li and Shiyu Zhang, et al.}

\begin{abstract}

Detectors with high coverage have direct and far-reaching benefits for road users in route planning and avoiding traffic congestion, but utilizing these data presents unique challenges including: the dynamic temporal correlation, and the dynamic spatial correlation caused by changes in road conditions. Although the existing work considers the significance of modeling with spatial-temporal correlation, what it has learned is still a static road network structure, which cannot reflect the dynamic changes of roads, and eventually loses much valuable potential information. To address these challenges, we propose DetectorNet enhanced by Transformer. Differs from previous studies, our model contains a Multi-view Temporal Attention module and a Dynamic Attention module, which focus on the long-distance and short-distance temporal correlation, and dynamic spatial correlation by dynamically updating the learned knowledge respectively, so as to make accurate prediction. In addition, the experimental results on two public datasets and the comparison results of four ablation experiments proves that the performance of DetectorNet is better than the eleven advanced baselines.

\end{abstract}



\maketitle

\section{Introduction}
	In the Intelligent Transportation System (ITS), detector data gradually occupy a pivotal position. According to PEMS\footnote{http://pems.dot.ca.gov/}, the highway system in all major metropolitan areas of California is covered with more than 39000 detectors. This report shows that detector is a common data-collection equipment on the road, and has great research value, especially for traffic prediction. 
	
	However, the traffic prediction problem based on detector network is confronted with many challenges. Specifically, it could be divided into the following points: (a) The dynamic spatial correlation.  (b) The dynamic temporal correlation. (c) Complex spatial-temporal correlation. Spatial correlation and temporal correlation are inseparable. In a nutshell, the traffic situation of one road may be closely related to the historical traffic situation of the surrounding roads.

	Potential research work can be used for solving the traffic prediction for the detector network. In early days, the task is simple viewed as the prediction of a multivariate time series. Therefore, time series models (e.g. ARIMA) that capture the periodicity of traffic data are widely used, but most of them cannot effectively model nonlinear time series and do not consider spatial correlation. Although grid-based methods~\cite{MDL} have achieved good results in city-level traffic forecast, the road network structure is naturally composed of Non-Euclidean data, which leads to a tough obstacle to applying these kinds of methods. In recent years, there have been some research works~\cite{MRA-BGCN} based on Graph Neural Networks (GNN) that consider the spatial-temporal correlation, but they merely concentrate on the modeling of road spatial relationships, while ignoring the dynamics of road spatial relationships. At the same time, although some efforts\cite{GraphWaveNet} have been made to solve the long-distance multi-step prediction, the effect needs to be improved.
	
	For the sake of overcoming the aforementioned problems and better tackling the traffic prediction problem based on the detector network, we lay our eyes upon the following two aspects: (1) Deeply consider the dynamics of road relations. It is not enough to consider the distance and spatial location between roads in isolation or learn a fixed relationship from traffic data, let alone it cannot reflect the change of road relationship. (2) Consider the temporal dependence of the road from multiple perspectives. Mining temporal correlation through historical observations of long-distance, medium-distance, and long-distance views respectively, has proved its effectiveness. But this is rarely considered on the monitor network. In conclusion, considering historical traffic conditions from multiple perspectives can better forecast future traffic conditions. On the basis of our thoughts and motivation, we propose an effective model named DetectorNet for graph-based traffic prediction. 

	Our main contributions are summarized as follows:
	\begin{itemize}
	    \item We emphasized the significance of dynamic spatial and temporal correlations, and proposed DetectorNet accordingly.
        \item We designed a multi-view temporal attention module and a dynamic spatial graph convolutional network, which respectively strengthen the learning of the temporal correlation of different views and the spatial correlation of different traffic conditions.
        \item We have proved the effectiveness of DetectorNet by comparing 11 baselines on two public traffic datasets.
	\end{itemize}
	
	\section{Problem Definition}\label{sec:pre}
	
	The urban detector network is regarded as a graph represented by $G=(V,E)$, where $V$ is the set of detector nodes and $E$ is the set of edges. The number of sensors denotes $|V|=N$, and the adjacency matrix of a detector graph refers to $A_{ij}\in{R^{N\times{N}}}$. At each time step $t$, detector graph $G$ has a graph signal matrix $X_{(t)}\in{R^{N\times{D}}}$, which initially represents the value recorded by each detector. Thus, given the history $P$ step graph signals, traffic prediction is to learn a function $f$ to predict the next $Q$ step graph signals. It is formulated as follows:
	\begin{equation}
		f(X_{(t-P+1):(t)},G)\rightarrow{X_{(t+1):(t+Q)}}
	\end{equation}
	where $X_{(t-P+1):(t)}\in{R^{P\times{N}\times{D}}}$ is historical observation and $X_{(t+1):(t+Q)}$ $\in{R^{Q\times{N}\times{D}}}$ is the prediction target.

	\section{METHODOLOGY OF DetecotrNet}\label{sec:methodlogy}



	\begin{figure*}[!t]
		\centering
		\includegraphics[width=1\textwidth]{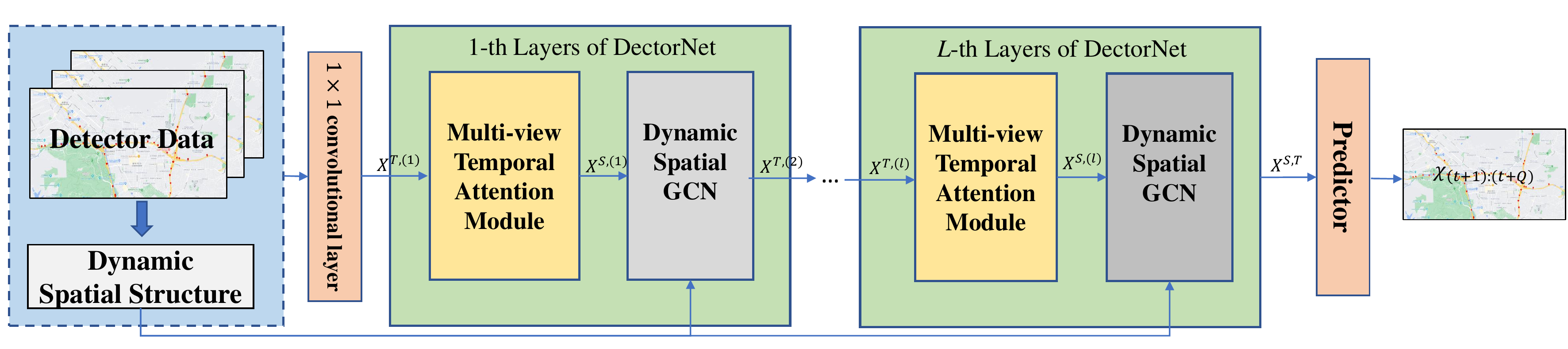}
		\caption{Overview of DetectorNet.}\label{overview}
	\end{figure*}
	
	\subsection{Multi-view Temporal Attention Module}
	\subsubsection{Data Preparation}
	To improve the feature expression, the dimension of original observation is increased by 1D-Convolution to obtain $X^{T,(1)}\in{\mathbb{R}^{N\times{P}\times{C_1}}}$. For clarity, we divide the input of the $l$-th layer $X^{T,(l)}$ into three view, namely short-distance view, medium-distance view and long-distance view. 
	\subsubsection{Multi-view Self Attention}
	For the observation of each perspective, we extract the features of dynamic temporal correlation separately based on the self attention mechanism, which dynamically adjust the attention coefficient according to the input. In each view, the feature at each time interact in pairs. To facilitate the specific description, we take short-distance view as an example.
	
	First, $X^{T,(l)}_{short}$ is mapped to three feature spaces: the \textbf{query} $Q^{T,(l)}_{short}$, the \textbf{key} $K^{T,(l)}_{short}$, and the \textbf{value} $V^{T,(l)}_{short}$. 

	We directly learn the correlation between the two time steps by:
	\begin{equation} \label{eq5}
		A^{T,(l)}_{short} = \text{Softmax}(\frac{Q^{T,(l)}_{short}(K^{T,(l)}_{short})^T}{\sqrt{d_{k}}})
	\end{equation}
	where $d_k\in{\mathbb{R}^{C_{l+1}}}$ is used to scale the result of the dot-product for a better training, and $\alpha_{ij}\in A^{T,(l)}_{short}$ represents the attention of time step $i$ to time step $j$.
	
	Then, we weight the temporal attention to the value, and obtain a new feature representation of a short-distance view:
	
	\begin{equation} 
		M^{T,(l)}_{short} = A^{T,(l)}_{short} V^{T,(l)}_{short} \in{\mathbb{R}^{N\times{m}\times{C_{l+1}}}}
	\end{equation}
	Similary, we calculate new representations of medium-distance and long-distance views: $M^{T,(l)}_{medium}$ and $M^{T,(l)}_{long}$.
	
	\subsubsection{Global Temporal Attention}
	
	We adopt the self-attention mechanism to model the relevance of all time steps. Unlike MTA, GTA doesn't need to divide the data and directly maps the input $X^{T,(l)}$ to the feature spaces of the query,the value and the key:
	\begin{equation} 
		\begin{split}
			Q^{T,(l)}_{global}=W^{g}_{xq}X^{T,(l)}, K^{T,(l)}_{global}=W^{g}_{xk}X^{T,(l)}, V^{T,(l)}_{global}=W^{g}_{xv}X^{T,(l)}
		\end{split}
	\end{equation}
	where $\{ W^{g}_{xq} ,W^{g}_{xk} ,W^{g}_{xv}\} \in{\mathbb{R}^{C_{l}\times{C_{l+1}}}}$ are learnable linear mapping matrices.
	
	Next, we use the scaled dot-product to simulate the pairwise interation of time steps, and obtain the global temporal attention matrix $A^{T,(l)}_{global}$. We formulate this process, which is equivalent to Equation \ref{eq5}.
	\begin{equation} 
		\begin{split}
			e^{T,(l)} &= \frac{Q^{S,(l)}_{short}(K^{T,(l)}_{short})^T}{\sqrt{d_{k}}} \in{\mathbb{R}^{P\times{P}}} \\
			\alpha_{ij} &= \frac{exp(e^{T,(l)}_{i,j})}{\sum_{k=1}^{P}exp(e^{T,(l)}_{i,k})},\alpha_{ij}\in{A^{T,(l)}_{global}}
		\end{split}
	\end{equation}
	
	Naturally, ${A^{T,(l)}_{global}}$ and $V^{T,(l)}_{global}$ are multipiled to get the global temporal features $M^{T,(l)}_{global}$.
	\subsubsection{Fusion}
	In this part, we merge multiple views and global temporal features. MTA gives the temporal information of three views (e.g. short-distance view). Through the concatenate operation, we get the following results: 
	\begin{equation} 
		M^{T,(l)}_{Multi-view} = Concat(M^{T,(l)}_{short}+M^{T,(l)}_{medium}+M^{T,(l)}_{short})
	\end{equation}
	where $M^{T,(l)}_{Multi-view}\in{\mathbb{R}^{N\times{P}\times{C_{l+1}}}}$ represent the result of multiple views. 
	
	We combine the residual connection and propose the fusion method as follow:
	\begin{equation} 
		\tilde{X}^{T,(l+1)} = M^{T,(l)}_{Multi-view} + \beta M^{T,(l)}_{global} + \gamma W_{res}X^{T,(l)}
	\end{equation}
	where $\beta$ and $\gamma$  are the weight coefficient, which can be learnable or set manually.
	
	Thus, we get the output of the $l$-th MTAM:
	\begin{equation} 
		X^{T,(l+1)} = f(ReLU{(W_{2} ReLU{(W_{1} \tilde{X}^{T,(l+1)})})}  + \tilde{X}^{T,(l+1)})
	\end{equation}
	where $f(x)$ represents the layer normalization operation, and $\{W_{1},W_{2}\}$ are the parameters of FCs.
	\subsection{Dynamic Spatial Graph Convolution Network}

	\begin{figure}
		\includegraphics[width=0.495\textwidth]{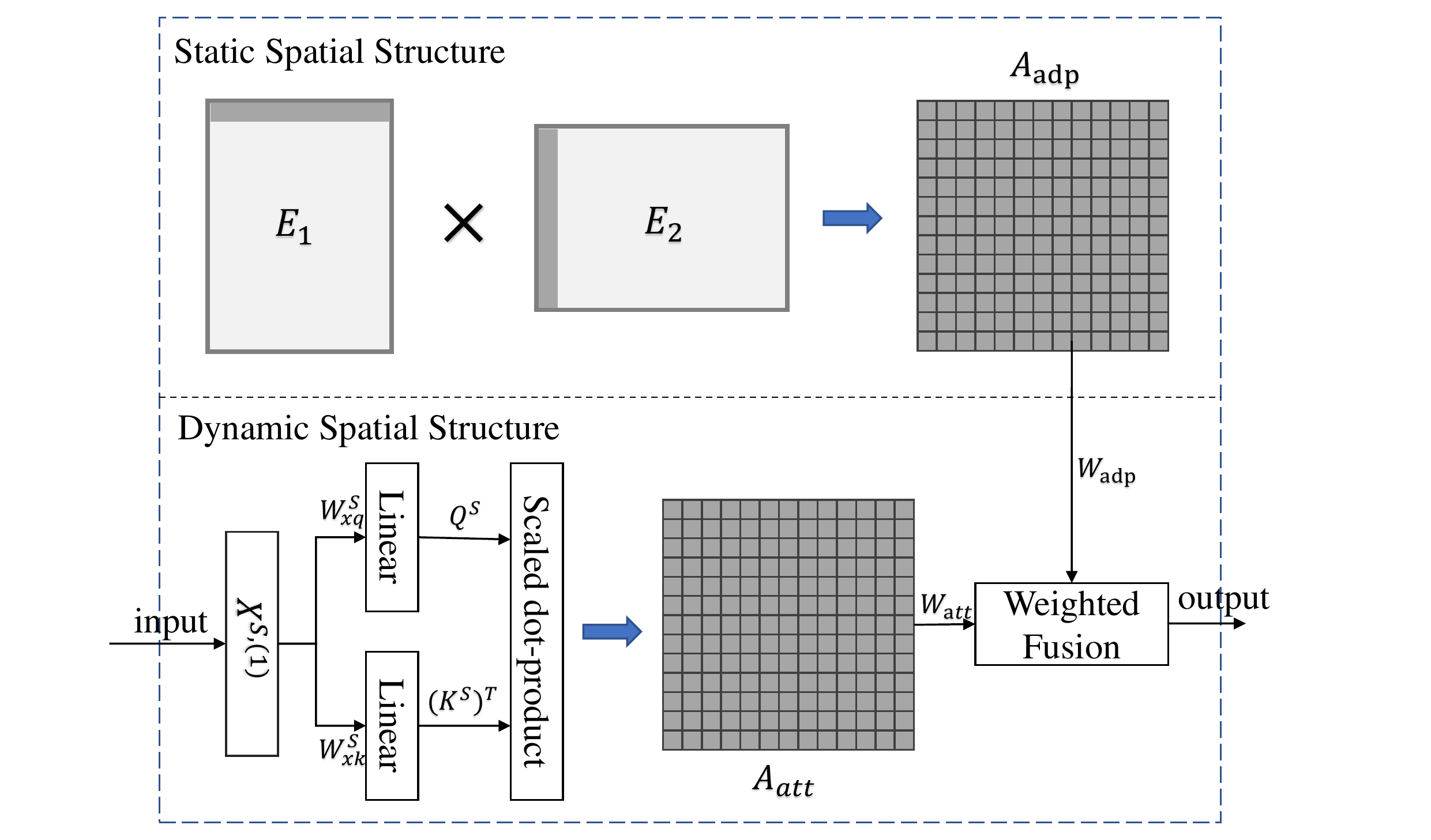}
		\caption{Dynamic Spatial Structure.}\label{dsgcn}
	\end{figure}

	\subsubsection{Dynamic Spatial Structure}
	
	
	We model the dynamic spatial relationship to reflect the dynamic change relationship of the road.

	Therefore, we need to learn useful information from the input data to capture this feature.
	For the input $X^{S,(l)}$ with temporal characteristics, we map it into query and key spaces in parallel, and get the interaction relationship between two roads through scaled dot-product.
	Next, using the dynamics of attention adjacency matrix and the structural features learned from adaptive matrix to obtain the dynamic spatial structure between roads (namely dynamic adjacency matrix) as follow:
	\begin{equation}  
		A_{dynamic} = Softmax(W_{att}\odot{A_{att}}+W_{adp}\odot{A_{adp}})
	\end{equation}
	where $\odot$ represents the Hadamard product, and $\{W_{att},W_{adp}\}$ are the learnable weighted parameters.
	
	\subsubsection{Graph Convolutional Network}
	We use the learned dynamic spatial structure to improve the process based on the Diffusion GCN, so as to capture dynamic spatial relations. This novel module is also named Dynamic Spatial GCN (DSGCN). Our modeling method is as follows:
	\begin{equation}  
		Z^{(l)}_{dynamic} = \sum_{k=0}^{K}{P^{k}_{f}X^{S,(l)}W_{k1}+P^{k}_{b}X^{S,(l)}W_{k2}+A^{k}_{dynamic}X^{S,(l)}W_{k3}}
	\end{equation}
	
	
	We eventually utilize FFN as a way to enhance the expression ability of Dynamic GCN. The specific formula is as follows:
	
	\begin{equation} 
		X^{S,(l+1)} = f(ReLU{(W_{2} ReLU{(W_{1}Z^{(l)}_{dynamic})})}  + Z^{(l)}_{dynamic})
	\end{equation}
	where $f(x)$ represents the layer normalization operation, and $\{W_{1},W_{2}\}$ are the parameters of FNN.
	
	\subsection{Predictor and Loss}
	In order to effectively utilize the extracted spatial and temporal features, we employ convolution as our predictor. In addition, for the sake of alleviating the loss of information, we use 2-layer convolution to fuse features step by step. The output of the last layer $X^{S,(L+1)}$ is deformed and rewritten as $X^{S,T}\in{\mathbb{R}^{N\times{Q}\times{C_{ST}}}}$,where $C_{ST}\times{Q}$ is equal to $c_{L+1}\times{P}$. Generally, we set $P=Q$. 
	So, the multi-steps prediction is
	\begin{equation} 
		\mathcal{X}_{(t+1):(t+Q)} = Conv(Conv(X^{S,T}_{L}))\in{\mathbb{R}^{N\times{Q}\times{c_{p}}}}
	\end{equation}
	Where $L$ represents the number of layers of DetectorNet, and $c_{p}$ is the length of the traffic volume to be predicted. Finally, we use MAE as the loss function to update parameters.

	\section{Experiment}\label{sec:experiment}

	\subsection{Datasets}
	\begin{table}[]
		\caption{Statistical specifics from METR-LA and PEMS-BAY. For instance, "Edges" represents the roads constructed by the connection of detectors.}
		\label{datasets}
		\resizebox{0.5\textwidth}{!}{%
			\begin{tabular}{@{}ccccccc@{}}
				\toprule
				Dataset  & Nodes & Edges & Time Samples & Sample Rate & Input Len. & Output Len. \\ \midrule
				METR-LA  & 207   & 1515  & 34272 &5min  & 12 &12    \\
				PEMS-BAY & 325   & 2369  & 52116  &5min &12  &12    \\ \bottomrule
			\end{tabular}
		}
	\end{table}

	We conduct a lot of experiments on two public traffic datasets namely PEMS-BAY and METR-LA. 
	Table ~\ref{datasets} shows a more concise introduction and statistical data of these two datasets.
	
	
	
	\subsection{Experimental Setup}
	
	\emph{Experimental Settings}. We utilize PyTorch to implement our model, DetectorNet. In our experiments, we set the number of DetectorNet to 2 with a 32-dimension hidden layer, $\beta$ and $\gamma$ are set to 1, and use graph convolution layer with a diffusion step = 2. Furthermore, the Adam optimizer is utilized to train our mode with a batch size of 64, and the initial learning rate is set to $0.001$, with a decay rate of 0.5 per 100 epochs. Besides, delay weight (i.e. L2 loss norm) is set to ${1e^{-5}}$. To alleviate overfitting, we set dropout to 0.3. Finally, we use three metrics (i.e. MAE, MAPE, RMSE) to evaluate the results.

	\begin{table}
		\caption{This table clearly shows the comparison between DetectorNet and the existing methods. From the statistical data, we can intuitively find that DetectorNet performs outstanding in comparison with other baselines.}
		\label{result}
		\setlength{\tabcolsep}{0.55mm}{\small
			\begin{tabular}{@{}cccccccc@{}}
				\toprule
				\multirow{2}{*}{Data} & \multirow{2}{*}{Model} &  \multicolumn{3}{c}{30min} & \multicolumn{3}{c}{60min} \\ \cmidrule(l){3-8} 
				&  & MAE & RMSE & MAPE & MAE & RMSE & MAPE  
				\\ \midrule
				\multirow{11}{*}{\rotatebox{90}{METR-LA}} 
				& HA  & 4.16 & 7.80 & \multicolumn{1}{c|}{13.00\%} & 4.16 & 7.80 & 13.00\% \\
				& ARIMA & 5.15 & 10.45 & \multicolumn{1}{c|}{12.70\%} & 6.9 & 13.23 & 17.40\% \\
				& FC-LSTM & 3.77 & 7.23 & \multicolumn{1}{c|}{10.90\%} & 4.37 & 8.69 & 13.20\% \\
				& WaveNet & 3.59 & 7.28 & \multicolumn{1}{c|}{10.25\%} & 4.45 & 8.93 & 13.62\% \\
				& DCRNN & 3.15 & 6.45 & \multicolumn{1}{c|}{8.80\%} & 3.6 & 7.6 & 10.50\% \\
				& ST-MetaNet & 3.10 & 6.28 & \multicolumn{1}{c|}{8.57\%} & 3.59 & 7.52 & 10.63\% \\
				& STGCN & 3.47 & 7.24 & \multicolumn{1}{c|}{9.57\%} & 4.59 & 9.4 & 12.70\% \\
				& MRA-BGCN\cite{MRA-BGCN} & 3.06 & 6.17 & \multicolumn{1}{c|}{8.30\%} & 3.49 & 7.30 & 10.00\% \\
				& GraphWaveNet\cite{GraphWaveNet} & 3.07 & 6.22 & \multicolumn{1}{c|}{{8.37}\%} & 3.53 & 7.37 & {10.01}\% \\
				& GMAN\cite{GMAN} & {3.07} & {6.34} & \multicolumn{1}{c|} {{8.35\%}} & \textbf{3.40} & {7.21} & {9.72\%} \\ 
				& MTGNN\cite{MTGNN} & \textbf{3.05} & 6.17 & \multicolumn{1}{c|}{8.19\%} & 3.49 & 7.23 & 9.87\% \\
				
				& DetectorNet & {3.06} & \textbf{6.08} & \multicolumn{1}{c|}{\textbf{8.12\%}} & \textbf{3.40} & \textbf{6.98} & \textbf{9.60\%} \\ 
				
				& DetecorNet w/o MTA & 3.10 & 6.23 & \multicolumn{1}{c|}{8.33\%} & 3.48 & 7.15 & 9.85\% \\
				& DetecorNet w/o GTA & 3.08 & 6.14 & \multicolumn{1}{c|}{8.48\%} & 3.46 & 7.00 & 9.87\% \\
				& DetecorNet w/o DA & 3.07 & 6.13 & \multicolumn{1}{c|}{8.29\%} & 3.45 & 7.07 & 9.78\% \\
				& DetecorNet w/o SA & 3.13 & 6.32 & \multicolumn{1}{c|}{8.37\%} & 3.52 & 7.24 & 10.09\% \\
				
				\midrule
				\multirow{11}{*}{\rotatebox{90}{PEMS-BAY}} 
				& HA & 2.88 & 5.59 & \multicolumn{1}{c|}{6.80\%} & 2.88 & 5.59 & 6.80\% \\
				& ARIMA & 2.33 & 4.76 & \multicolumn{1}{c|}{5.40\%} & 3.38 & 6.5 & 8.30\% \\
				& FC-LSTM & 2.2 & 4.55 & \multicolumn{1}{c|}{5.20\%} & 2.37 & 4.96 & 5.70\% \\
				& WaveNet & 1.83 & 4.21 & \multicolumn{1}{c|}{4.16\%} & 2.35 & 5.43 & 5.87\% \\
				& DCRNN & 1.74 & 3.97 & \multicolumn{1}{c|}{3.90\%} & 2.07 & 4.74 & 4.90\% \\
				& ST-MetaNet & 1.76 & 4.02 & \multicolumn{1}{c|}{4.00\%} & 2.20 & 5.06 & 5.45\% \\
				& STGCN & 1.81 & 4.27 & \multicolumn{1}{c|}{4.17\%} & 2.49 & 5.69 & 5.79\% \\
				& MRA-BGCN\cite{MRA-BGCN} & {1.61} & 3.67 & \multicolumn{1}{c|}{3.80\%} & 1.91 & 4.46 & {4.60\%} \\
				& GraphWaveNet\cite{GraphWaveNet} & 1.63 & 3.7 & \multicolumn{1}{c|}{3.67\%} & 1.95 & 4.52 & 4.62\% \\
				& GMAN\cite{GMAN} & {1.62} & {3.72} & \multicolumn{1}{c|} {{3.63\%}} & {1.86} & {4.32} & {4.31\%} \\ 
				& MTGNN\cite{MTGNN} & 1.65 & 3.74 & \multicolumn{1}{c|}{3.69\%} & 1.94 & 4.49 & 4.53\% \\

				& DetectorNet & \textbf{1.57} & \textbf{3.54} & \multicolumn{1}{c|}{3.56\%} & 
				\textbf{1.80} & 4.26 & \textbf{4.19\%} \\ 
				
				& DetecorNet w/o MTA & 1.59 & 3.59 & \multicolumn{1}{c|}{3.59\%} & 1.84 & 4.31 & 4.39\% \\
				& DetecorNet w/o GTA & 1.61 & 3.63 & \multicolumn{1}{c|}{3.65\%} & 1.86 & 4.35 & 4.41\% \\
				& DetecorNet w/o DA & 1.61 & 3.58 & \multicolumn{1}{c|}{\textbf{3.50\%}} & 1.83 & 4.23 & 4.15\% \\
				& DetecorNet w/o SA & 1.61 & 3.57 & \multicolumn{1}{c|}{3.65\%} & 1.84 & \textbf{4.20} & 4.28\% \\
				
				\bottomrule
			\end{tabular}
		}
	\end{table}

	\subsection{Performance Comparison}
	We validate DetectorNet on two datasets, and the result is reported in Table ~\ref{result}. 
Furthermore, we have the following findings:
	
	The method based on Graph Neural Network (GNN) is suitable for the spatial-temporal prediction of the detector network. From the view of the experimental results, the GNN based method (e.g. GraphWaveNet, DetectorNet) performs obviously better than the traditional time series based method (e.g. WaveNet). We firmly believe the reason is that the traditional models ignore the importance of spatial correlation and treat the temporal information of each node in isolation. 
	
	Self-Attention mechanism can effectively model long-distance sequences. Obviously, DetectorNet based on Self-Attention mechanism has a significant advantage in remote multi-step prediction, which is far superior to RNN based and Temporal Convolution Networck (TCN) based methods. Indeed, it must be acknowledged that DetectorNet still has room for improvement in short-term prediction.

	\section{Conclusion}
	
	In this paper, a novel model named DetectorNet is proposed to achieve a better addressing of spatial-temporal graph-based traffic prediction. By utilizing a Multi-view Temporal Attention modules and other modules, DetectorNet has the ability to not only consider the original static informaton, but also capture the dynamically varying correlation of the road structure, eventually make accurate prediction on traffic flow. Experimental results conducted on two open datasets verify the superiority of DetectorNet when compared to other eleven baselines.

	\begin{acks}
    This work was supported by National Natural Science Foundation of China (61602354, 61876138), Natural Science Foundation of Shaanxi Province (2019JM-227), National Research Foundation of Korea(NRF) grant funded by the Korea government(MSIT). (No. 2019R1A2C2084257), and the MSIT(Ministry of Science and ICT), Korea, under the Grand Information Technology Research Center support program(IITP-2021-2020-0-01462) supervised by the IITP(Institute for Information \text{\&} communications Technology Planning \text{\&} Evaluation).
	\end{acks}

	\bibliographystyle{ACM-Reference-Format}
\end{document}